\documentclass[10pt,twocolumn,letterpaper]{article}
\usepackage{cvpr}
\usepackage{times}
\usepackage{epsfig}
\usepackage{graphicx}
\usepackage{amsmath}
\usepackage{amssymb}
\usepackage{color}
\usepackage{booktabs}
\usepackage[breaklinks=true,bookmarks=false]{hyperref}
\usepackage[margin=4pt,font=footnotesize,labelfont=bf,labelsep=endash,tableposition=top]{caption}

\cvprfinalcopy

\def\cvprPaperID{} % *** Enter the CVPR Paper ID here

\begin{document}

%%%%%%%%% TITLE
\title{PF-Net: Point Fractal Network for 3D Point Cloud Completion}

%\author{
%{\large
%Zitian Huang$^{1}$, ~Yikuan Yu$^{1,2}$, ~Jiawen Xu$^{1}$, ~Feng Ni$^{2}$, ~Xinyi Le$^{1*}$}\\

%{\large
%${^1}$Shanghai Jiao Tong University ~~~~${^2}$SenseTime}\\
%\small E-mail: $\tt Zitian.huang@sjtu.edu.cn $
%}
\author{Zitian Huang$^{1}$, ~Yikuan Yu$^{1,2}$, ~Jiawen Xu$^{1}$, ~Feng Ni$^{2}$, ~Xinyi Le$^{1*}$\\
${^1}$ Shanghai Jiao Tong University,~~~~${^2}$ SenseTime\\
{\tt\small Zitian.huang,lexinyi@sjtu.edu.cn}}

\maketitle
%\thispagestyle{empty}

%%%%%%%%% ABSTRACT
\begin{abstract}
    In this paper, we propose a Point Fractal Network (PF-Net), a novel learning-based approach for precise and high-fidelity point cloud completion. Unlike existing point cloud completion networks, which generate the overall shape of the point cloud from the incomplete point cloud and always change existing points and encounter noise and geometrical loss, PF-Net preserves the spatial arrangements of the incomplete point cloud and can figure out the detailed geometrical structure of the missing region(s) in the prediction. To succeed at this task, PF-Net estimates the missing point cloud hierarchically by utilizing a feature-points-based multi-scale generating network. Further, we add up multi-stage completion loss and adversarial loss to generate more realistic missing region(s). The adversarial loss can better tackle multiple modes in the prediction. Our experiments demonstrate the effectiveness of our method for several challenging point cloud completion tasks.

\end{abstract}

%%%%%%%%% BODY TEXT
\section{Introduction}
3D vision has become one of the current research topics recently. Among various types of 3D data description, point cloud has been widely used in 3D data processing due to its small data size but more delicate presenting ability. Real-world point cloud data is often captured by using laser scanners, stereo cameras, or low-cost RGB-D scanners. However, due to occlusion, light reflection, transparency of surface material, and limitations of sensor resolution and viewing angle, it will cause a loss of geometric and semantic information, resulting in incomplete point clouds. Therefore, it is  an essential task to repair incomplete point clouds for further applications.
\begin{figure}[ht]
\centering  %插入的图片居中表示
\includegraphics[width=1\linewidth]{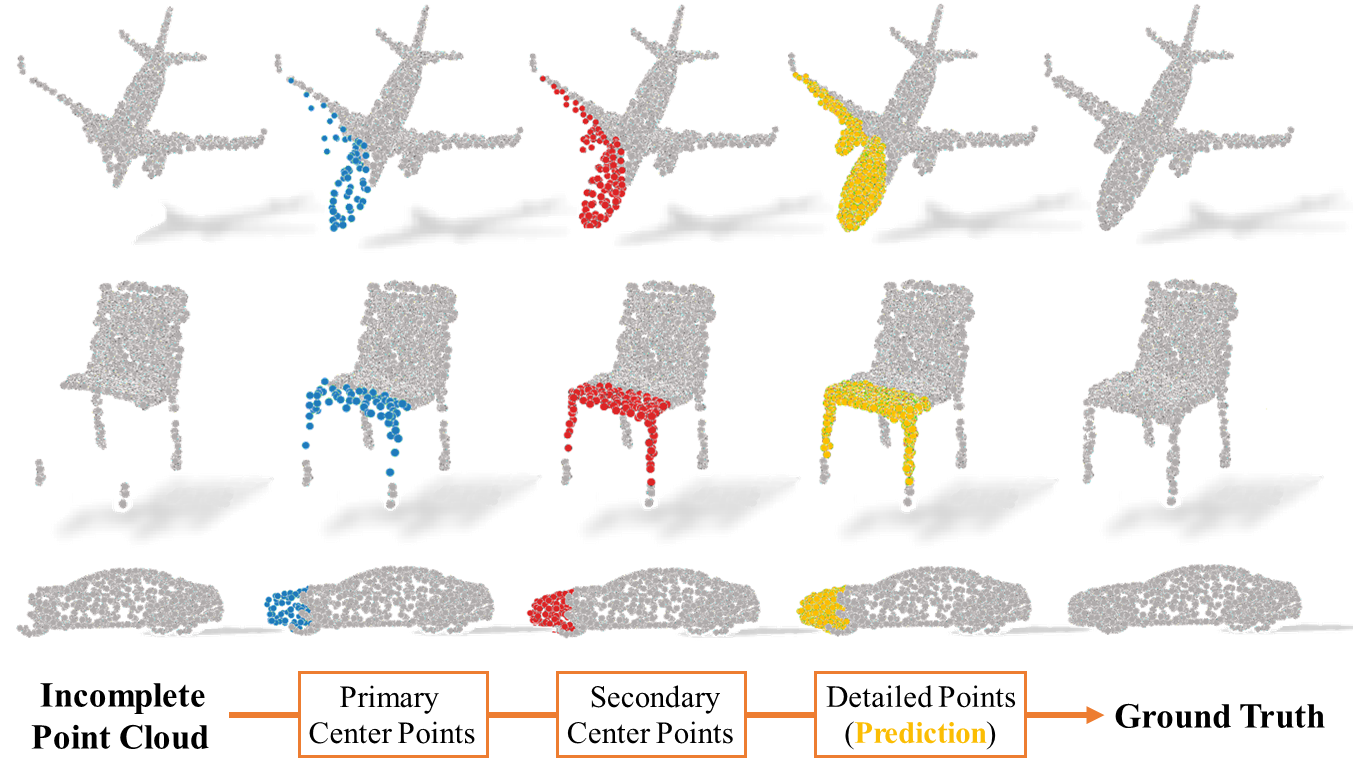}
\caption{Our PF-Net takes the incomplete point cloud as input and directly predicts the missing point cloud. The display of multi-scale predictions (indicated by blue, red and yellow) are made by layers from different depths of Point Pyramid Decoder (PPD). The low resolution prediction serves as the center point to generate high resolution prediction, leading to a coarse-to-fine prediction.} %图片的名称
%\fbox{\rule{0pt}{2in}
%\rule{.9\linewidth}{0pt}}
\label{Fig.1}   %标签，用作引用
\end{figure}
Since the 3D point cloud is unstructured and unordered, the majority of the deep-learning-based methods for dealing with 3D data transform point cloud to collections of images(e.r, views) or regularly voxel-based representations of 3D data. However, multi-view and voxel-based representation\cite{liu2019relation-shape,feng2019hypergraph,yu2018multi-view,he2018triplet-center,kanezaki2018rotationnet:,zhou2018voxelnet:} leads to unnecessarily voluminous and  limits the output resolution of voxels.\cite{qi2016pointnet:,yuan2018pcn:}

Thanks to the advent of PointNet \cite{qi2016pointnet:}, learning-based architectures are capable of directly operating on point cloud data. L-GAN \cite{achlioptas2018learning} introduces the first deep-learning-based network for point cloud completion by utilizing an Encoder-Decoder framework. PCN \cite{yuan2018pcn:} combines the advantages of L-GAN \cite{achlioptas2018learning} and FoldingNet \cite{yang2017foldingnet:} which is specialized on repairing incomplete point clouds. Recently, RL-GAN-Net \cite{sarmad2019rl-gan-net:} proposes a reinforcement learning agent controlled GAN for reducing prediction time of point cloud completion.

Previous works for completion take incomplete point cloud as input and aim to output the overall complete models. They pay more attention to learning the general character of a genus/category but not the local details of a specific object. Therefore, they may change the position of known points, and encounter genus-wise distortions \cite{yang2017foldingnet:}, which causes noise and detailed geometrical loss.  For example, in Fig. \ref{Fig.4}(3), the reconstruction of a chair in previous works miss the existing cross-bar (under the chair), and fail to generate the hollow back. The reason may be that Auto-Encoders tend to average multiple ``chairs'' in the prediction. Noticing that most chairs in the training set have fully filled back without cross-bar, as is shown in Fig. \ref{Fig.1}, the previous networks thus are more likely to generate normal chairs with full filled back other than ``special'' chairs with hollow back and cross-bars in Fig. \ref{Fig.4}(3).
\begin{figure*}[htb]
\centering  %插入的图片居中表示
\includegraphics[width=1\linewidth]{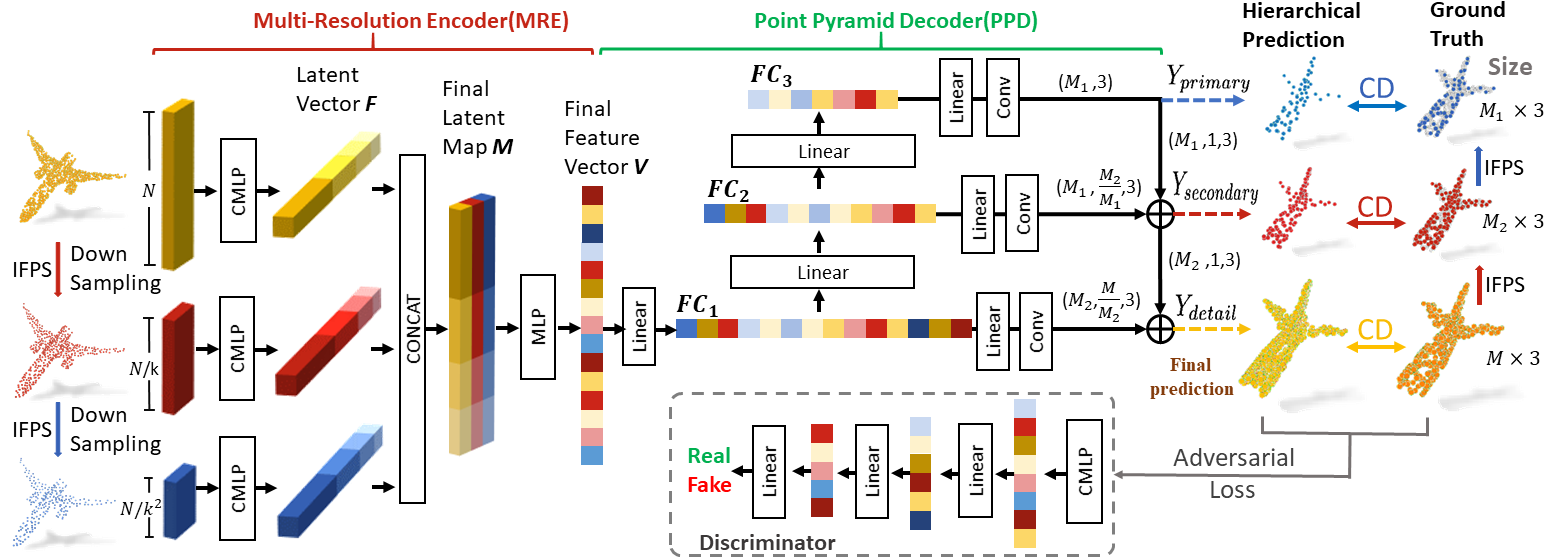}
\caption{{\bf The architecture of PF-Net.} The input of PF-Net is incomplete point cloud and it predicts the missing point cloud by utilizing a multi-scale generating network PPD. IFPS (Section 3.1) samples the input (indicated by yellow) into different scales (indicated by red and blue). Three scales of prediction are indicated by blue, red and yellow (final prediction) in the figure. CMLP (Section 3.2) is our feature extractor. CD (Chamfer Distance in Section 3.4) is applied to evaluate the difference between the prediction and the ground truth.}  %图片的名称
%\fbox{\rule{0pt}{2in}
%\rule{.9\linewidth}{0pt}}
\label{Fig.2}   %标签，用作引用
\end{figure*}
In this work, we present the unsupervised Point Fractal Network (PF-Net) for repairing the incomplete point cloud. Different from the existing Auto-Encoder architectures, the main features of PF-Net can be summarized as follows:
%keep the invariance of the original part

(1) To preserve the spatial arrangements of the original part, we take the partial point cloud as input and only output the missing part of the point cloud instead of the whole object. This architecture has two benefits: Firstly, it retains the geometric features of the original point cloud after repairing. Secondly, it helps the network focus on perceiving the location and structure of missing parts. The main difficulty of such kind of prediction is that partial known points and points to be inferred are merely related in semantic level and completely different in spatial arrangements.

(2) For better feature extraction of the specific partial point cloud, we first propose a novel Multi-Resolution Encoder (MRE) to extract multi-layer features from the partial point cloud and its low resolution feature points by utilizing a new feature extractor Combined Multi-Layer Perception (CMLP). These multi-scale features contain both local and global features, also low-level and high-level features, enhancing the ability of the network to extract semantic and geometric information.

(3) To tackle the genus-wise distortions problem \cite{yang2017foldingnet:} and generate fine-grained missing region(s), we design a Point Pyramid Decoder (PPD) to generate the missing point cloud hierarchically. PPD is a multi-scale generating network based on feature points. PPD will predict primary, secondary, and detailed points from layers of different depth. Primary points and secondary points will try to match their corresponding feature points and serve as the skeleton center points to propagate the overall geometry information to the final detailed points. Fig. \ref{Fig.1} visualizes the process of this hierarchical prediction, which shows excellent performance on generating a high-quality missing region point cloud and restoring the original detailed shape.

In addition, we propose a multi-stage completion loss to guide the prediction of our method. Multi-stage completion loss helps the network pay more attention to the feature points. To further alleviate the distortions problem caused by Auto-Encoder framework, inspired by a renowned 2D context encoder \cite{pathak2016context}, we reduce this burden in the loss function by jointly train our PF-Net to minimize both multi-stage completion loss and adversarial loss. Adversarial loss optimizes our network that enable it to select a specific point cloud from multiple modes.

\section{Related Work}

%-------------------------------------------------------------------------
\subsection{Shape Completion and Deep Learning}

Currently, 3D shape completion methods are mainly based on the 3D voxel grid and point cloud. For voxel grid-based algorithm, architectures such as 3D-RecGAN \cite{yang20173d}, 3D-Encoder-Predictor Networks (3D-EPN) \cite{dai2017shape}, and hybrid framework of 3D-ED-GAN and LRCN \cite{wang2017shape} have been developed to realize the goal of repairing incomplete input data. However, voxel-based methods are restricted by its resolution because the computation cost increases dramatically with resolution.

PointNet \cite{qi2016pointnet:} solves the problem of processing unordered point sets. Thus, algorithms which can directly conduct shape completion task on unordered point have been significantly developed \cite{li2018point,Han2017High,wu2016learning,yu2018pu-net:}. L-GAN \cite{achlioptas2018learning} introduces the first deep generative model for the point cloud. Though L-GAN \cite{achlioptas2018learning} is capable of conducting shape completion tasks to some extent, its architecture is not primarily built to do shape completion task, and thus the performance is not considered ideal. FoldingNet \cite{yang2017foldingnet:} introduces a new decoding operation named Folding, serving as a 2D-to-3D mapping. Later, Point Completion Network (PCN) \cite{yuan2018pcn:} proposes the first learning-based architecture focusing on shape completion task. PCN \cite{yuan2018pcn:} applies the Folding operation \cite{yang2017foldingnet:} to approximate a relatively smooth surface and conduct shape completion. Recently, a reinforcement learning agent controlled GAN based network (RL-GAN-Net) \cite{sarmad2019rl-gan-net:} is invented for real-time point cloud completion. An RL agent is used in \cite{sarmad2019rl-gan-net:} to avoid complex optimization and accelerate prediction process, but it does not focus on enhancing prediction accuracy of the points. 3D point capsule networks \cite{sabour2017dynamic,zhao20183d,jaiswal2018capsulegan:,wang2018an} surpasses the performance of other methods and becomes the state-of-the-art Auto-Encoder processing point cloud, especially in the domain of point cloud reconstruction.

%-------------------------------------------------------------------------
\subsection{Context Encoder and Feature Pyramid Network}

Context encoder \cite{pathak2016context} processes images with missing region and trains a convolutional neural network to regress the missing pixel values. This architecture is formed by an encoder capturing the unspoiled content of the image and a decoder to predict the missing image content. In addition, it jointly train reconstruction and adversarial loss for semantic inpainting.

Feature Pyramid Network (FPN) \cite{Lin2017Feature} is an efficient method to solve the problem of multi-scale feature extraction. Thanks to the architecture of FPN \cite{Lin2017Feature}, the final feature map is the fusion of both semantically-rich and locally-rich features \cite{chen2019hybrid,tian2019fcos:,pang2019libra,cai2018cascade,Kirillov2019Panoptic}. It enhances the geometric and semantic information contained in the final feature map. FPN \cite{Lin2017Feature} partially inspires the decoder in our network.

%------------------------------------------------------------------------
\section{Method}
In this section, we will introduce our PF-Net, which predicts the missing region of the point cloud from its incomplete known configuration. Fig. \ref{Fig.2} shows the complete architecture of our PF-Net. The overall architecture of our PF-Net is composed of three fundamental building blocks, named Multi-Resolution Encoder (MRE), Point Pyramid Decoder (PPD), and Discriminator Network.

%The Multi-Resolution Encoder takes three different resolution of partial point cloud as input and outputs feature map which contains latent vectors in different dimension. Feature Pyramid Decoder then takes in the feature map and produces primary center points, secondary center points and the detailed point cloud of the missing region. The final prediction and the ground truth will be fed into Discriminator. Discriminator tries to distinguish synthesized point cloud from the ground truth and optimize MRE and FPD to predict the correct shape. Note that, all our three blocks are trained together in one stage and we only use MRE and FPD to generate prediction after the training process.

%------------------------------------------------------------------------
\subsection{Feature Points Sampling}
\begin{figure}[ht]
\centering  %插入的图片居中表示
\includegraphics[width=0.8\linewidth]{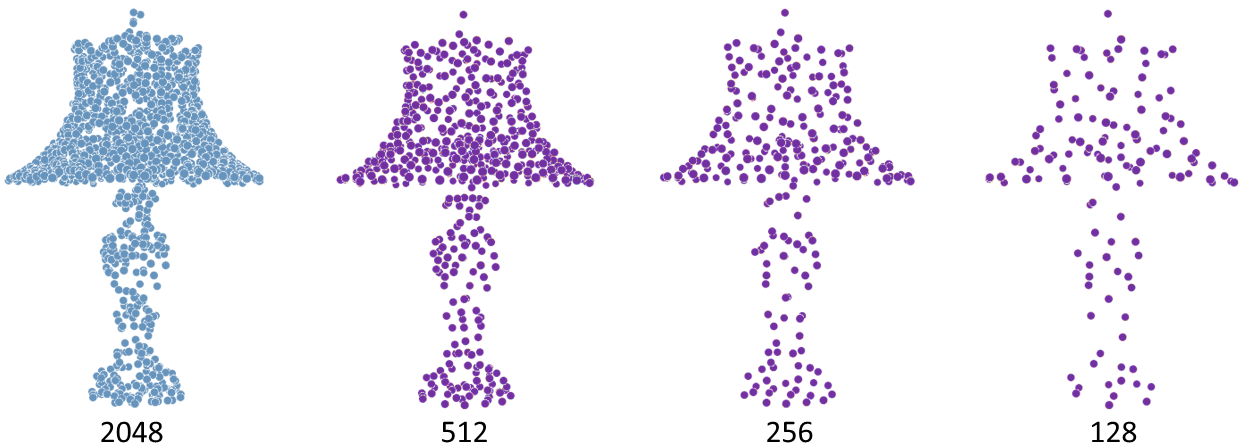}
\caption{{\bf Results of IFPS in different resolutions.} The blue lamp has 2048 points. Three purple lamps are sampled from the blue lamp with 512, 256, and 128 points, respectively.} %图片的名称
%\fbox{\rule{0pt}{2in}
%\rule{.9\linewidth}{0pt}}
\label{Fig.3}   %标签，用作引用
\end{figure}

In a point cloud, we can only take a small number of points to characterize the shape, which are defined as feature points. In other words, feature points map the skeleton of a point cloud\cite{birdal2017a}. To extract feature points from a point cloud, we use iterative farthest point sampling (IFPS), which is a sampling strategy applied in Pointnet++ \cite{qi2017pointnet++:} to get a set of skeleton points. IFPS can represent the distribution of the entire point sets better compared to random sampling, and it is more efficient than CNNs \cite{qi2017pointnet++:}. In Fig. \ref{Fig.3}, we visualize the effect of IFPS. Even if we only extract 6.25\% of the points, these points can still describe the configuration of the lamp and have a similar density distribution to the original lamp.
%------------------------------------------------------------------------
\subsection{Multi-Resolution Encoder}
\begin{figure}[h]
\centering  %插入的图片居中表示
\includegraphics[width=1\linewidth]{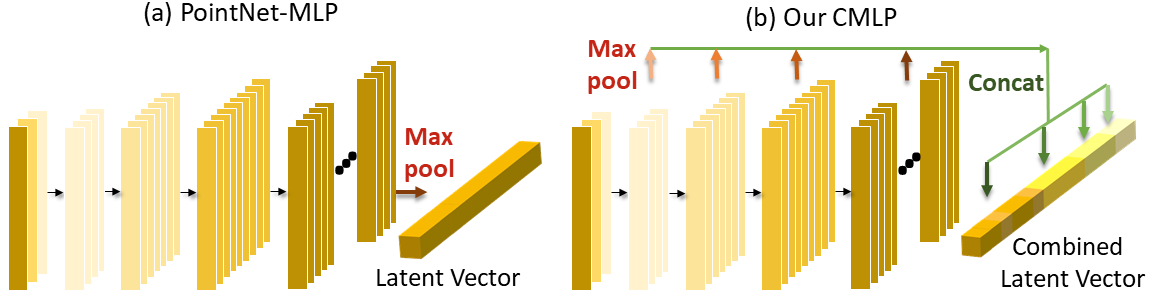}
\caption{{\bf Comparison of two different feature extractors of 3D point encoders.} PointNet-MLP only uses the final single layer to extract
the latent vector. Our CMLP maps the combined latent vector by
concatenating multiple dimensional latent vectors derived from
the last four layers. It can make better use of the high-level and low-level features. }%图片的名称
%\fbox{\rule{0pt}{2in}
%\rule{.9\linewidth}{0pt}}
\label{Fig.4}   %标签，用作引用
\end{figure}

\noindent\textbf{CMLP}
We first introduce the feature extractor of our MRE, named as Combined Multi-Layer Perception (CMLP). In most of the previous work, the feature extractor of their encoder is Multi-Layer Perception (MLP). We refer to this as PointNet-MLP, shown in Fig. \ref{Fig.4}(a). This method maps each point into different dimensions and extracts the maximum value from the final $K$ dimensions to form a global latent vector. However, it does not make good use of low-level and mid-level features that contain rich local information. Also, L-GAN \cite{achlioptas2018learning} and PointNet \cite{qi2016pointnet:}  show that the performance of this feature extractor is strongly affected by the dimension of the Maxpooling layer, i.e., $K$. In CMLP, we also use MLP to encode each point into multiple dimensions $[64-128-256-512-1024]$. Different from MLP, we maxpool the output of the last four layers of MLP to obtain multiple dimensional feature vector f$_{i}$, where size f$_{i}$ := $128,~ 256,~ 512,~ 1024,$  for $i = 1,...,4$. All the f$_{i}$ are then concatenated, forming the combined latent vector  $\bf{F}$. Size \textbf{F} := $1920$, and it contains both low-level and high-level feature information.

The input to our Multi-Resolution Encoder is an $N \times 3$ unordered point cloud. We downsample the input point cloud to obtain two more scales in different resolutions (Size: $\frac{N}{k} \times 3$ and $\frac{N}{k^2} \times 3$) by using IFPS. We use this data-dependent way to get the feature points of the input point sets, helping the encoder focus on those more representative points. Three independent CMLPs will be used to map those three scales into three individual combined latent vector \textbf{F$_{i}$}, for $i = 1,2,3$. Each vector represents the feature extracted from a certain resolution of the point cloud. All the \textbf{F$_{i}$} are then concatenated, forming a latent feature map \textbf{M} in the size of $1920 \times 3$ ($i.e.$, three vectors each in the size of 1920). We then use MLP [3 - 1] to integrate the latent feature map into a final feature vector \textbf{V}. Size of \textbf{V}=$1920$.

%------------------------------------------------------------------------
\subsection{Point Pyramid Decoder}
\begin{figure}[ht]
\centering  %插入的图片居中表示
\includegraphics[width=1\linewidth]{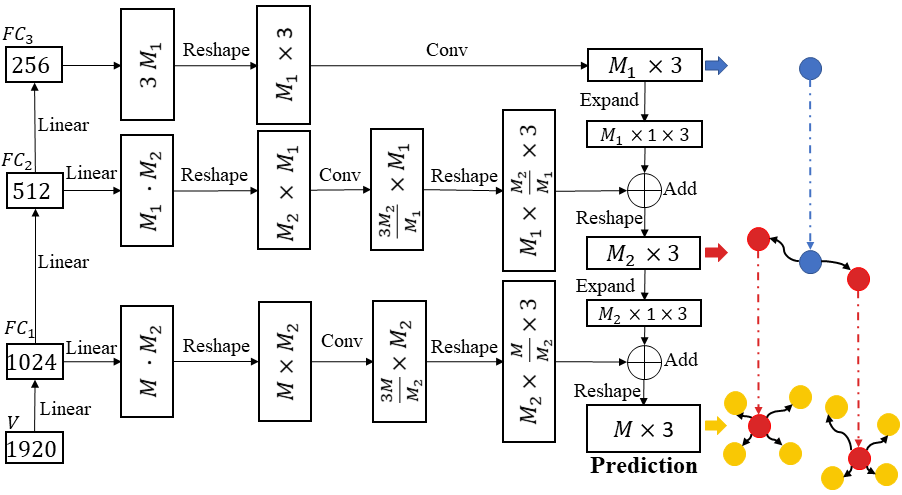}
\caption{{\bf Details of Point Pyramid Decoder}. PPD gives outputs of three joint layers in ascending scales, which are similar both in semantic and geometric level. Intuitively, the architecture of PPD reflects the concept of ``Fractal geometry" in mathematics.}
%"Fractal geometry" defines that a fragmentary geometry can be divided into several parts, each of which is the reduced shape of the whole.} % 图片的名称
%\fbox{\rule{0pt}{2in}
%\rule{.9\linewidth}{0pt}}
\label{Fig.5}   %标签，用作引用
\end{figure}

The decoder takes final feature vector \textbf{V} as input and aims to output $M \times 3$ point cloud, which represents the shape of the missing region. The baseline of our PPD is fully-connected decoder \cite{achlioptas2018learning}. A fully-connected decoder is good at predicting the global geometry of point cloud. Nevertheless, it always causes loss of local geometric information since it only uses the final layer to predict the shape. Previous works combine fully-connected decoder with a folding-based decoder to reinforce the local geometry of the predicted shape. However, as shown in \cite{yang2017foldingnet:}, a folding-based decoder is not good at handling genus-wise distortions \cite{yang2017foldingnet:} and retaining the original detailed geometry if the original surface is relatively complicated. To overcome the limitation, we design our PPD as a hierarchical structure based on feature points, which is inspired by FPN \cite{Lin2017Feature}. The details of the PPD are shown in Fig. \ref{Fig.5}. We first obtain three feature layers $\mathbf{FC}_{i}$ (size $\mathbf{FC}_{i}$ :=1024, 512, 256 neurons, for $i = 1,~2,~3$) by passing \textbf{V} through fully-connected layers. Each feature layer is responsible for predicting point cloud in different resolutions. The primary center points $Y_{\rm primary}$ will be predicted from the deepest $\mathbf{FC}_{3}$, having the size of ${M_1} \times 3$. Then, $\mathbf{FC}_{2}$ will be used to predict the relative coordinates of secondary center points $Y_{\rm secondary}$. Each point in $Y_{\rm primary}$ will serve as center point to generate $\frac{M_2}{M_1}$ points of $Y_{\rm primary}$. We implement this process using ``Expand'' and ``Add'' operations. Thus, size of $Y_{\rm secondary}$ is  ${M_2} \times 3$. Detailed points $Y_{\rm detail}$ is the final prediction of PPD. The generation of $Y_{\rm detail}$ is similar to $Y_{\rm secondary}$ shown in Fig. \ref{Fig.5}. Size $Y_{\rm detail}$: ${M} \times 3$. Meanwhile,  $Y_{\rm detail}$ will try to match the feature points sampled from G.T.. Thanks to this multi-scale generating architecture, high-level features will affect the expression of low-level features, and low resolution feature points can propagate the local geometry information to the high resolution prediction. Our experiments show that the prediction of PPD has fewer distortions and can retain the detailed geometry of the original missing point cloud.

%------------------------------------------------------------------------
\subsection{Loss Function}
%------------------------------------------------------------------------
The loss function of our PF-Net consists of two parts: multi-stage completion loss and adversarial loss. Completion loss measures the difference between the ground truth of missing point cloud $Y_{\rm gt}$ and the predicted point cloud. Adversarial loss tries to make the prediction seem more real by optimizing the MRE and PPD. The size of $Y_{\rm gt}$ is ${M} \times 3$ which is the same as $Y_{\rm detail}$.

\noindent\textbf{Multi-stage Completion Loss}
Fan \cite{Fan2016A} has proposed two permutation-invariant metrics to compare unordered point cloud which are Chamfer Distance (CD) and Earth Mover's Distance (EMD). In this work, we choose Chamfer Distance as our completion loss since it is differentiable and more efficient to compute compared to EMD.
{\small
\begin{align}
\label{Eq.1}
    d_{\rm CD}(S_{1},S_{2}) = \frac{1}{S_{1}} \sum\limits_{x \in S_{1}} \underset{y \in S_{2}}{\rm min} \Vert x-y \Vert_2^2 + \frac{1}{S_{2}} \sum\limits_{y \in S_{2}} \underset{x \in S_{1}}{\rm min} \Vert y-x \Vert_2^2
\end{align}}
CD in (\ref{Eq.1}) measures the average nearest squared distance between predicted point cloud $S_{1}$ and the ground truth point cloud $S_{2}$. Since the Point Pyramid Decoder will predict three point cloud in different resolution, our multi-stage completion loss consists of three terms in (\ref{Eq.2}), $d_{\rm CD_1}$, $d_{\rm CD_2}$, and $d_{\rm CD_3}$, weighted by hyperparameter $\alpha$. The first term calculates the squared distance between detailed points $Y_{\rm detail}$ and the ground truth of missing region $Y_{\rm gt}$. The second and third term calculate the squared distance between primary center points $Y_{\rm primary}$, secondary center points $Y_{\rm secondary}$ and the subsampled ground truth $Y'_{\rm gt}$, $Y''_{\rm gt}$ respectively. The subsampled ground truth $Y'_{\rm gt}$ and $Y''_{\rm gt}$ have the same size as $Y_{\rm primary}$ (${M_1} \times 3$) and $Y_{\rm secondary}$ (${M_2} \times 3$), respectively. We obtain $Y'_{\rm gt}$ and $Y''_{\rm gt}$ from $Y_{\rm gt}$ by applying IFPS. They are the feature points of the missing region. The design of the multi-stage completion loss increases the proportion of feature points, leading to better focus on feature points.
\begin{align}
\label{Eq.2}
    L_{\rm com} &= d_{\rm CD1}(Y_{\rm detail},Y_{\rm gt}) +
   \alpha\ d_{\rm CD2}(Y_{\rm primary},Y'_{\rm gt}) \notag\\&+
    2\alpha\ d_{CD3}(Y_{\rm secondary},Y''_{\rm gt})
\end{align}

\noindent\textbf{Adversarial Loss}
The adversarial loss in this work is inspired by Generative Adversarial Network (GAN) \cite{goodfellow2014generative}. We first define $\rm F() := \rm PPD(\rm MRE())$, where MRE is Multi-Resolution Encoder and PPD is Point Pyramid Decoder. $F : \mathcal{X}\rightarrow\mathcal{Y'}$ will map the partial input $\mathcal{X}$ into predicted missing region $\mathcal{Y'}$. Then Discriminator ($\rm D()$) tries to distinguish the predicted missing region $\mathcal{Y'}$ and the real missing region $\mathcal{Y}$. The discriminator is a classification network with similar structure as CMLP. The specific structure has serial MLP layers $[64 - 64 - 128 - 256]$ and we maxpool the response from the last three layers to obtain feature vector $f_{i}$, where size $f_{i}$ := 64, 128, 256, for $i$ = 1, 2, 3. Concatenate them into a latent vector $\rm F$. The size of $\rm F$ is 448. $\rm F$ will be passed through fully-connected layers [256,128,16,1] followed by Sigmoid-classifier to obtain predicted value. We now define the adversarial loss:
\begin{align}
\label{Eq.3}
    L_{\rm adv} = \sum\limits_{1\leq i\leq S} {\rm log}(D({y}_i))+\sum\limits_{1\leq j\leq S} {\rm log}(1-D(F(x_i)))
\end{align}
where $x_i\in\mathcal{X},y_i\in\mathcal{Y}$,$~i=1,\dots,S$. $S$ is the dataset size of $\mathcal{X},\mathcal{Y}$.  Both $F$ and $D$ are optimized jointly by using alternating Adam when training.

\noindent\textbf{Joint Loss}
The joint loss of our network is defined as:
\begin{equation}
      L =\lambda_{\rm com}L_{\rm com} + \lambda_{\rm adv}L_{\rm adv}
\end{equation}
$\lambda_{com}$ and $\lambda_{adv}$ are the weight of completion loss and adversarial loss which satisfy the followed condition: $\lambda_{com} + \lambda_{adv} = 1$.

%------------------------------------------------------------------------
\section{Experiments}
%In this section, we first introduce a multi-category dataset we created to train our model.Then, we compare our method against the existing point cloud completion networks and the state-of-the-art point cloud reconstruction networks in different views. Finally,  we will display our completion results.

%-------------------------------------------------------------------------
\subsection{Data Generation and Implementation detail}
%-------------------------------------------------------------------------
To train our model, we use 13 categories of different objects in the benchmark dataset Shapenet-Part \cite{yi2016a}. The total number of shapes sums to 14473 (11705 for training and 2768 for testing). All input point cloud data is centered at the origin, and their coordinates are normalized to [-1,1]. The ground-truth point cloud data is created by sampling 2048 points uniformly on each shape. The incomplete point cloud data is generated by randomly selecting a viewpoint as a center in multiple viewpoints and removing points within a certain radius from complete data. We control the radius to get a different amount of missing points. When comparing our method against other methods, we set the incomplete point cloud missing 25\% of the original data for train and test.

We implement our network on PyTorch. All the three building blocks are trained by using ADAM optimizer alternately with an initial learning rate of 0.0001 and a batch size of 36. We employ batch normalization (BN) and RELU activation units at MRE and Discriminator but only use RELU activation units (except for the last layer) at PPD. In MRE, we set $k=2$. In PPD, we set $M_1=64$, $M_2=128$ based on the numbers of points of each shape. And we only change $M$ to control the size of the final prediction.
\begin{table*}[]
\centering
\begin{tabular}{cccccc}
\toprule[2pt]
Category   & LGAN-AE\cite{achlioptas2018learning}         & PCN\cite{yuan2018pcn:}             & 3D-Capsule\cite{zhao20183d}      & PF-Net(vanilla)& PF-Net        \\ \hline
\hline
Airplane   & 0.856 $/$ 0.722 & 0.800 $/$ 0.800 & 0.826 $/$ 0.881 & 0.284 $/$  \textbf{0.231} & \textbf{0.263} $/$0.238 \\
Bag        & 3.102 $/$ 2.994 & 2.954 $/$ 3.063 & 3.228 $/$ 2.722 & 0.927 $/$ 0.934           & \textbf{0.926} $/$ \textbf{0.772} \\
Cap        & 3.530 $/$ 2.823 & 3.466 $/$ 2.674 & 3.439 $/$ 2.844 & 1.308 $/$ \textbf{1.027}  & \textbf{1.226} $/$ 1.169 \\
Car        & 2.232 $/$ 1.687 & 2.324 $/$ 1.738 & 2.503 $/$ 1.913 & 0.616 $/$ 0.431           & \textbf{0.599} $/$ \textbf{0.424} \\
Chair      & 1.541 $/$ 1.473 & 1.592 $/$ 1.538 & 1.678 $/$ 1.563 & \textbf{0.472} $/$ \textbf{0.420} & 0.487 $/$ 0.427 \\
%Earphone   & 2.899 $/$ 3.183 & 2.789 $/$ 3.528 & 3.337 $/$ 2.395 & \textbf{0.997} $/$ 1.802  & 1.475 $/$ \textbf{1.676} \\
Guitar     & 0.394 $/$ 0.354 & 0.367 $/$ 0.406 & 0.298 $/$ 0.461 & \textbf{0.097} $/$ 0.094  & 0.108 $/$ \textbf{0.091} \\
%Knife      & 0.479 $/$ 0.461 & 0.412 $/$ 0.436 & 0.440 $/$ 0.466 & 0.131 $/$ \textbf{0.107}  & \textbf{0.106} $/$ 0.117 \\
Lamp       & 3.181 $/$ 1.918 & 2.757 $/$ 2.003 & 3.271 $/$ 1.912 & 1.041 $/$ \textbf{0.616}  & \textbf{1.037} $/$ 0.640 \\
Laptop     & 1.206 $/$ 1.030 & 1.191 $/$ 1.155 & 1.276 $/$ 1.254 & 0.309 $/$ \textbf{0.244}  & \textbf{0.301} $/$ 0.245 \\
Motorbike  & 1.828 $/$ 1.455 & 1.699 $/$ 1.459 & 1.591 $/$ 1.664 & 0.524 $/$ 0.414           & \textbf{0.522} $/$ \textbf{0.389} \\
Mug        & 2.732 $/$ 2.946 & 2.893 $/$ 2.821 & 3.086 $/$ 2.961 & 0.793 $/$ 0.776           & \textbf{0.745} $/$ \textbf{0.739} \\
Pistol     & 1.113 $/$ 0.967 & 0.968 $/$ 0.958 & 1.089 $/$ 1.086 & 0.270 $/$ \textbf{0.237}  & \textbf{0.252} $/$ 0.244 \\
%Rocket     & 0.812 $/$ 0.648 & 0.682 $/$ 0.648 & 0.855 $/$ 0.789 & 0.239 $/$ \textbf{0.166}  &\textbf{0.222} $/$ 0.172 \\
Skateboard & 0.887 $/$ 1.02  & 0.816 $/$ 1.206 & 0.897 $/$ 1.262 & 0.289 $/$ 0.288           & \textbf{0.225} $/$ \textbf{0.172} \\
Table      & 1.694 $/$ 1.601 & 1.604 $/$ 1.790 & 1.870 $/$ 1.749 & \textbf{0.505} $/$ 0.417  & 0.525 $/$ \textbf{0.404} \\ \hline
Mean       & 1.869 $/$ 1.615 & 1.802 $/$ 1.662 & 1.927 $/$ 1.713 & 0.572 $/$ 0.471 & \textbf{0.555} $/$ \textbf{0.458} \\
\toprule[2pt]
\end{tabular}
\caption{\textbf{Point cloud completion results of overall point cloud}. The training data consists of 13 categories of different objects \cite{yi2016a}. The numbers shown are [Pred $\rightarrow$ GT error $/$ GT $\rightarrow$ Pred error], scaled by 1000. We compute the mean values across all categories and show them in the last row of the table. PF-Net (vanilla) is PF-Net without Discriminator.}
\label{table1}
\end{table*}

\begin{table*}[]
\centering
\begin{tabular}{cccccc}
\toprule[2pt]
Category   & LGAN-AE\cite{achlioptas2018learning}         & PCN\cite{yuan2018pcn:}             & 3D-Capsule\cite{zhao20183d}       & PF-Net(
vanilla) & PF-Net       \\ \hline
\hline
Airplane   & 3.357 \textbf{$/$} 1.130 & 5.060 $/$ 1.243 & 2.676 $/$ 1.401 & 1.197 $/$ \textbf{1.006} & \textbf{1.091} $/$ 1.070 \\
Bag        & 5.707 $/$ 5.303 & \textbf{3.251} $/$ 4.314 & 5.228 $/$ 4.202 & 3.946 $/$ 4.054 & 3.929 $/$ \textbf{3.768} \\
Cap        & 8.968 $/$ 4.608 & 7.015 $/$ \textbf{4.240} & 11.04 $/$ 4.739 & 5.519 $/$ 4.470 & \textbf{5.290} $/$ 4.800 \\
Car        & 4.531 $/$ 2.518 & 2.741 $/$ 2.123 & 5.944 $/$ 3.508 & 2.537 $/$ 1.848          & \textbf{2.489} $/$ \textbf{1.839} \\
Chair      & 7.359 $/$ 2.339 & 3.952 $/$ 2.301 & 3.049 $/$ 2.207 & \textbf{1.998} $/$ 1.828 & 2.074 $/$ \textbf{1.824} \\
%Earphone   & 7.272 $/$ 4.846 & 8.505 $/$ 8.194 & 7.635 $/$ \textbf{3.743}                   & \textbf{4.612} $/$ 7.623 & 6.300 $/$ 6.715 \\
Guitar     & 0.838 $/$ 0.536 & 1.419 $/$ 0.689 & 0.625 $/$ 0.662 & \textbf{0.435} $/$ 0.435 & 0.456 $/$ \textbf{0.429} \\
%Knife      & 1.157 $/$ 0.528 & 1.457 $/$ 0.742 & 0.948 $/$ 0.688 & 0.725 $/$ \textbf{0.519} & \textbf{0.472} $/$ 0.534 \\
Lamp       & 8.464 $/$ 3.627 & 11.61 $/$ 7.139 & 9.912 $/$ 5.847 & 5.252 $/$ \textbf{3.059} & \textbf{5.122} $/$ 3.460 \\
Laptop     & 7.649 $/$ 1.413 & 3.070 $/$ 1.422 & 2.129 $/$ 1.733 & 1.291 $/$ 1.013          & \textbf{1.247} $/$ \textbf{0.997} \\
Motorbike  & 4.914 $/$ 2.036 & 4.962 $/$ 1.922 & 8.617 $/$ 2.708 & 2.229 $/$ 1.876          & \textbf{2.206} $/$ \textbf{1.775} \\
Mug        & 6.139 $/$ 4.735 & 3.590 $/$ 3.591 & 5.155 $/$ 5.168 & 3.228 $/$ 3.332          & \textbf{3.138} $/$ \textbf{3.238} \\
Pistol     & 3.944 $/$ 1.424 & 4.484 $/$ 1.414 & 5.980 $/$ 1.782 & 1.267 $/$ \textbf{1.012} & \textbf{1.122} $/$ 1.055 \\
%Rocket     & 8.803 $/$ 1.001 & 5.177 $/$ 1.005 & 12.57 $/$ 1.171 & 1.040 $/$ \textbf{0.734} & \textbf{0.933} $/$ 0.744 \\
Skateboard & 5.613 $/$ 1.683 & 3.025 $/$ 1.740 & 11.49 $/$ 2.044 & 1.198 $/$ \textbf{1.257} & \textbf{1.136} $/$ 1.337 \\
Table      & 2.658 $/$ 2.484 & 2.503 $/$ 2.452 & 3.929 $/$ 3.098 & \textbf{2.184} $/$ \textbf{1.928} & 2.235 $/$ 1.934 \\ \hline
Mean       & 5.395 $/$ 2.603 & 4.360 $/$ 2.661 & 5.829 $/$ 3.008 & 2.483 $/$ \textbf{2.086} & \textbf{2.426} $/$ 2.117 \\
\toprule[2pt]
\end{tabular}
\caption{\textbf{Point cloud completion results of the missing point cloud}. The numbers shown are [Pred $\rightarrow$ GT error $/$ GT $\rightarrow$ Pred error], scaled by 1000. In this table, We compute those two metrics in the missing region of point cloud.}
\label{table2}
\end{table*}

%-------------------------------------------------------------------------
\subsection{Unsupervised Point Cloud Completion Results}
%-------------------------------------------------------------------------

In this subsection, we compare our method against several representative baselines that operate directly on a 3D point cloud, including L-GAN \cite{lin2018learning}, PCN\cite{yuan2018pcn:}, 3D point-Capsule Networks \cite{zhao20183d}.
%L-GAN uses only fully-connected layers in decoder. PCN is based on FoldingNet which is a more complex method using on point processing. 3D Point Capsule Network is the top performer on many point reconstruction tasks thanks to its dynamic routing scheme and the reinforcement of FoldingNet.
Since all the existing methods are trained in different datasets, we train and test them in the same dataset in order to evaluate them quantificationally. It should be noted that all the methods are trained in an unsupervised way, which means that label information will not be provided. To evaluate the methods mentioned above, we use the evaluation metric by \cite{gadelha2018multiresolution,lin2018learning}. It contains two indexes: Pred $\rightarrow$ GT (prediction to ground truth) error and GT $\rightarrow$ Pred (ground truth to prediction) error. Pred $\rightarrow$ GT error computes the average squared distance from each point in prediction to its closest in ground truth. It measures how difference the prediction is from the ground truth. GT $\rightarrow$ Pred error computes the average square distance from each point in  the ground truth to its closest in prediction. which indicates the degree of the ground truth surface being covered by the shape of the prediction. We first compute Pred $\rightarrow$ GT error and GT $\rightarrow$ Pred error on the whole complete point cloud by concatenating the prediction of our network and the input partial point cloud.

Table \ref{table1} shows the result. Our method outperforms other methods in all categories on both Pred $\rightarrow$ GT and  GT $\rightarrow$ Pred error. Note that the error of the overall complete point cloud comes from two parts: the prediction error of the missing region and the change of the original partial shape. Since our method takes the partial shape as input and only output the missing region, it does not change the original partial shape. To make sure that our evaluation is reasonable, we also compute Pred $\rightarrow$ GT error and GT $\rightarrow$ Pred error on the missing region. Table \ref{table2} shows the results. %of completion only in the missing region for each method.
Our Method (PF-Net and PF-Net (vanilla)) outperforms existing methods in 12 out of 13 categories on both Pred $\rightarrow$ GT error and GT $\rightarrow$ Pred error. Furthermore, in terms of the mean of all 13 categories, our method has considerable advantages on both metrics. The result in Table \ref{table1} and Table \ref{table2} indicate that our method can generate more high-precision point cloud with less distortions both in the overall point cloud and the missing region point cloud.

\begin{figure*}[htb]
\centering  %插入的图片居中表示
\includegraphics[width=1\linewidth]{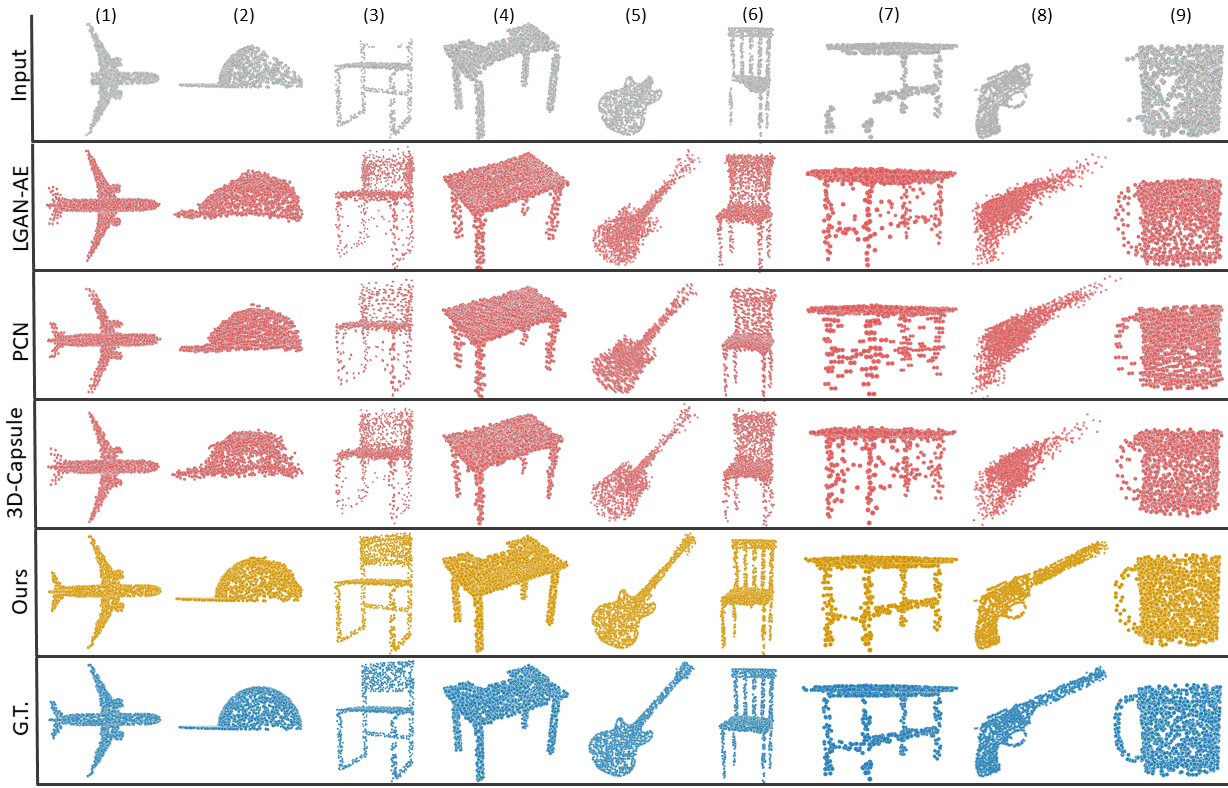}
\caption{{\bf The comparison of completion results between other methods and our network.} From top to bottom: Input, L-GAN \cite{achlioptas2018learning}, PCN \cite{yuan2018pcn:}, 3D-Capsule \cite{zhao20183d}, our method, and the ground truth. In (3)(7), the outputs of other methods lose the geometric structure of the cross-bar under the chair and the table. In (3)(4)(6), other methods mistakenly fill in the hollow part of the G.T. ( i.e., (3)(6), the back of the chair; (4), the space under the tabletop is mistakenly filled). In (8), others fail to generate a reasonable shape. Correspondingly, ours can generate the missing point cloud accurately with less distortions and high-level restoration. More results of ours are shown in Supplement.}  %图片的名称
%\fbox{\rule{0pt}{2in}
%\rule{.9\linewidth}{0pt}
\label{Fig.6}   %标签，用作引用
\end{figure*}

In Fig. \ref{Fig.6} we visualize the output point cloud generated by the methods mentioned above. All of them come from the test set. Compared to other methods, the prediction of our PF-Net presents strong spatial continuity with a high level of restoration and less genus-wise distortions. \cite{yang2017foldingnet:}.
%-------------------------------------------------------------------------
\subsection{Quantitative Evaluations of PF-Net}
%-------------------------------------------------------------------------

\noindent\textbf{Analysis of Discriminator}
The function of the Discriminator is to distinguish the predicted shape from the real profile of the missing region and optimize our network to generate a more ``realistic'' configuration. Table \ref{table1} and Table \ref{table2} show the result of PF-Net without Discriminator. Compared to PF-Net without Discriminator, PF-Net with Discriminator outperforms in 10 out of 13 categories on the Pred $\rightarrow$  GT error both in Table \ref{table1} and Table \ref{table2}. Furthermore, PF-Net has a considerable margin in terms of mean on the Pred $\rightarrow$  GT error both in Table \ref{table1} and Table \ref{table2}. The results indicate that PF-Net can help minimize Pred $ \rightarrow $ GT error. As is mentioned above, Pred $ \rightarrow $ GT error measures how difference the prediction is from the ground truth. Thus, the Discriminator enables PF-Net to generate the point cloud that is more similar to the ground truth.

\noindent\textbf{Analysis of MRE and PPD}
To demonstrate the effectiveness of CMLP, we train CMLP and other extractors which follow the same linear classification model on ModelNet40 \cite{wu20153d}and evaluate their overall classification accuracy on the test set. Results are shown in Table \ref{table3}. PointNet-MLP has the same parameter as PointNet without Transform-Net \cite{qi2016pointnet:}. PointNet-CMLP replaces MLP of PointNet-MLP by CMLP. Our-CMLP shows the best performance. The result also indicates that CMLP has a better understanding of semantic information.
\begin{table}[h]
\resizebox{\columnwidth}{!}
{
\begin{tabular}{cccc}
\toprule[1pt]
     & PointNet-MLP & PointNet-CMLP & Our-CMLP \\ \hline
Acc.($\%$) & 87.2         & 87.9          & 88.9      \\
\toprule[1pt]
\end{tabular}
}
\caption{Classification accuracy results of PointNet-MLP, PointNet-CMLP, and Our-CMLP on ModelNet40.}
\label{table3}
\end{table}

In order to assess the effectiveness of Multi-Resolution Encoder(MRE) and Point Pyramid Decoder(PPD), we further compare our PF-Net(vanilla) with three different baselines: single-resolution MLP, single-resolution CMLP, and multi-resolution CMLP. All the above methods take the incomplete point cloud as input and output the missing point cloud. We conduct this experiment on the models in ``Chair'' and ``Table'' class, which are two largest categories in the dataset we generated to discuss. ``Chair'' has $2658$ shapes for training and $704$ shapes for testing. ``Table'' has   $3835$ shapes for training and $848$ shapes for testing. Single-resolution MLP consists of 5 layers MLP [64,128,256,512,1024] and 2 linear layers [1024,1536], each layer followed by BN and ReLU activation (i.e we find that BN is beneficial in this method). Single-resolution CMLP has the same structure as single-resolution MLP but MLP is changed to CMLP. Multi-resolution CMLP use multi-resolution encoder followed by 2 linear layers [1024, 1536]. We compute their Pred$ \rightarrow$ GT error and GT $\rightarrow$ Pred error of the missing point cloud. Results are shown in Table \ref{table4}. Note that, in both two categories, CMLP and MR-CMLP are superior to MLP in Pred $\rightarrow$ GT error and GT $\rightarrow$ Pred error. Their performance boost is between 3\% to 10\% compared to MLP. However, after using PF-Net, the Pred $\rightarrow$ GT error and GT $\rightarrow$ Pred error are reduced significantly. This is what we expected. Since CMLP and MR-CMLP strengthen the ability of encoder to extract geometrical and semantic information, but their decoders only focus on generating the overall shape of the missing point cloud.

\begin{figure}[h]
\centering  %插入的图片居中表示
\includegraphics[width=0.85\linewidth]{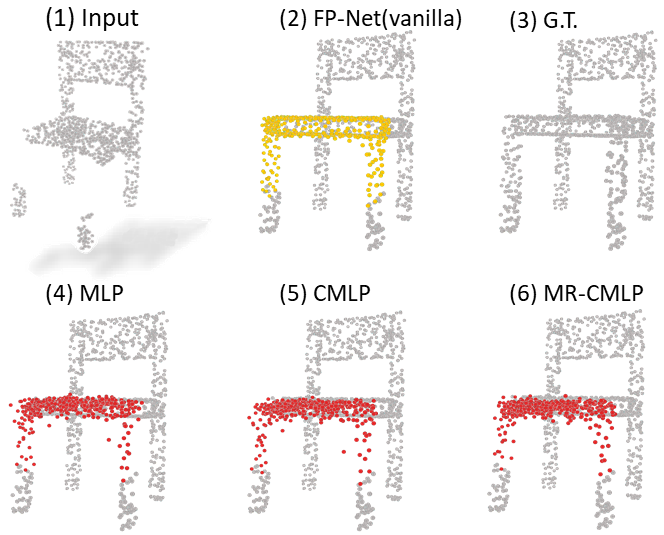}
\caption{\textbf{A chair predicted by MLP, CMLP , MR-CMLP, and PF-Net(vanilla)}. PF-Net(vanilla) is our PF-Net without Discriminator. Yellow and red points represent the prediction of PF-Net(vanilla) and others. It can be found that the middle of the chair in the G.T. is hollow. The prediction of MLP, CMLP, and MR-CMLP all lose this detailed structure. Our PF-Net(vanilla) retains this structure, and the prediction has less noise and distortions. } %图片的名称
%\fbox{\rule{0pt}{2in}
%\rule{.9\linewidth}{0pt}}
\label{Fig.7}   %标签，用作引用
\end{figure}

However, the generation process of PPD is smoother. It can focus on both the overall shape and the detailed feature points. Fig. \ref{Fig.7} depicts one example of the ``chair'' class in the test set.

%it is worth mentioning that our in-painting model can retain the initial shape of incomplete point cloud     not obvious in the image
%by raising the proportion of feature points in the loss function.
\begin{table}[!htbp]
\resizebox{\columnwidth}{!}
{
\begin{tabular}{ccccc}
\toprule[1pt]
Category & MLP         & CMLP        & MR-CMLP     & PF-Net(vanilla)                \\ \hline
Chair & 2.692 $/$ 2.384 & 2.587 $/$ 2.306 & 2.521 $/$ 2.275 & \textbf{1.772} $/$ \textbf{1.432} \\ \hline
Table & 2.429 $/$ 2.734 & 2.250 $/$ 2.693 & 2.299 $/$ 2.696 & \textbf{1.891} $/$ \textbf{1.664} \\ \hline
\toprule[1pt]
\end{tabular}
}
\caption{[Pred $\rightarrow$ GT error $/$ GT $\rightarrow$ Pred error] given by MLP, CMLP, MR-CMLP and our PF-Net(vanilla) of the missing point cloud.}
\label{table4}
\end{table}

\begin{figure}[h]
\centering  %插入的图片居中表示
\includegraphics[width=1\linewidth]{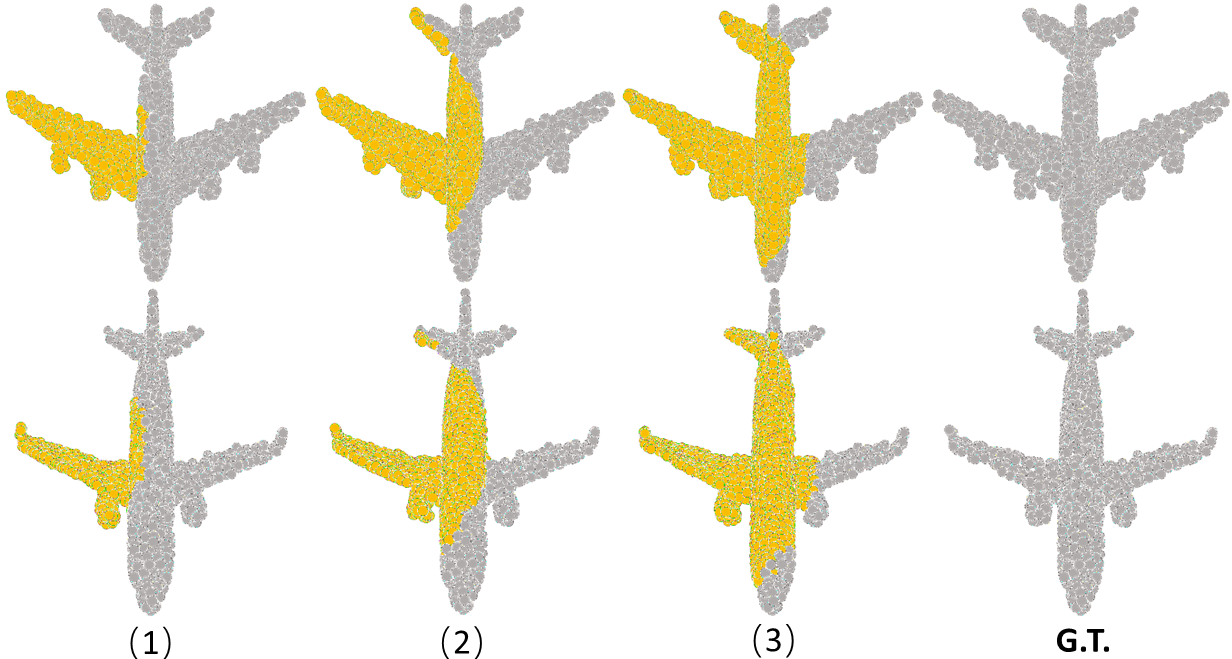}
\caption{\textbf{Examples of repairing results when the input has different extent of incompletion}. (1), (2), (3) lose 25\%, 50\%, 75\% points  of the original point cloud respectively. Yellow represents the prediction. Grey denotes the undamaged point cloud.} %图片的名称
%\fbox{\rule{0pt}{2in}
%\rule{.9\linewidth}{0pt}}
\label{Fig.8}   %标签，用作引用
\end{figure}

\begin{figure}[h]
\centering  %插入的图片居中表示
\includegraphics[width=0.95\linewidth]{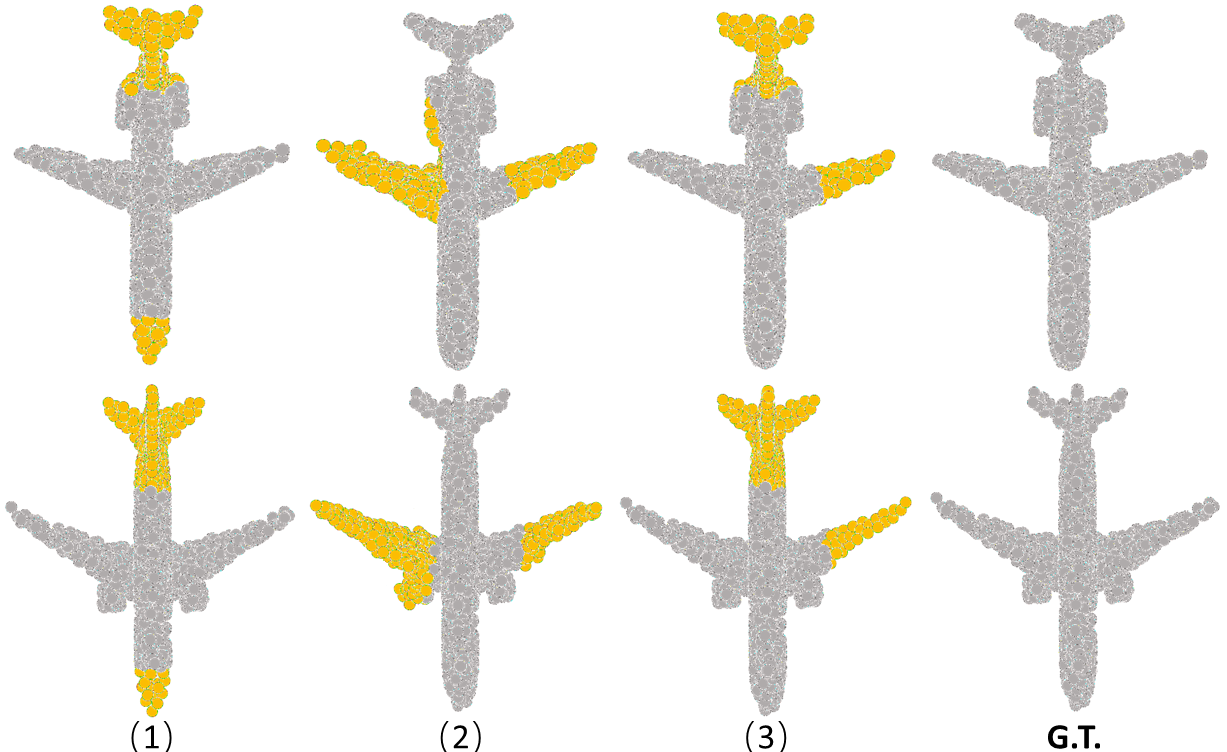}
\caption{Repairing results when the original point cloud loses different numbers of points in two random positions. Yellow represents the prediction. Grey denotes the undamaged point cloud.} %图片的名称
%\fbox{\rule{0pt}{2in}
%\rule{.9\linewidth}{0pt}}
\label{Fig.9}   %标签，用作引用
\end{figure}

\noindent\textbf{Robustness Test}
We conduct all the robustness test experiments on the ``Airplane'' class. In the first robustness test, we change $M$ in the PPD to control the number of the output points of our network and train it to repairing incomplete shape with different extent of incompletion, respectively. The experiment results are shown in Table \ref{table5}. 25\%, 50\% and 75\% mean that three incomplete inputs lose 25\%, 50\% and 75\% of the original points respectively compared to the ground truth. Note that the Pred $\rightarrow$ GT error and GT $\rightarrow$ Pred error of three partial inputs are basically the same, which implies that our network has strong robustness when dealing with incomplete inputs with different extent of missing. Fig. \ref{Fig.8} shows the performance of our network in the test set. Our network accurately ``identified'' different typed of airplane and retain the geometric details of the original point cloud even in the case of large-scale incompletion.
In the second robustness test, we train our network to complete the partial shape that lose more than one pieces of points in different positions.  One of the example in test set is shown in Fig. \ref{Fig.9}. It can be found that PF-Net still predict the correct missing point cloud in the right places.

\begin{table}[h]
\resizebox{\columnwidth}{!}
{
\begin{tabular}{cccc}
\toprule[1pt]
Missing Ratio      & 25\%        & 50\%        & 75\%        \\ \hline
PF-Net & 0.727 $/$ 0.542 & 0.727 $/$ 0.520 & 0.748 $/$ 0.623 \\
\toprule[1pt]
\end{tabular}
}
\caption{[Pred $\rightarrow$ GT error $/$ GT $\rightarrow$ Pred error] given by our PF-Net of the missing point cloud. The incomplete point cloud lose 25\%, 50\%, 75\% respectively compared to the original point cloud.}
\label{table5}
\end{table}

\section{Conclusion}
We have presented a new approach to shape completion using a partial point cloud only to generate the missing point cloud. Our architecture enables the network to generate the target point cloud with both rich semantic profile and detailed character while retaining the existing contour.
Our method outperforms the existing methods focusing on shape completion of a point cloud. Noticing that if the training dataset is as large enough, our method has the chance to repair any complex random point cloud delicately. Hence, we can foresee that if this method is applied deeply, the accuracy of 3D recognition can be dramatically improved, bringing new possibilities to the research of autonomous vehicles and 3D reconstruction.

\appendix
\section{Appendix}
\begin{figure*}[h]
\centering  %插入的图片居中表示
\includegraphics[width=1\linewidth]{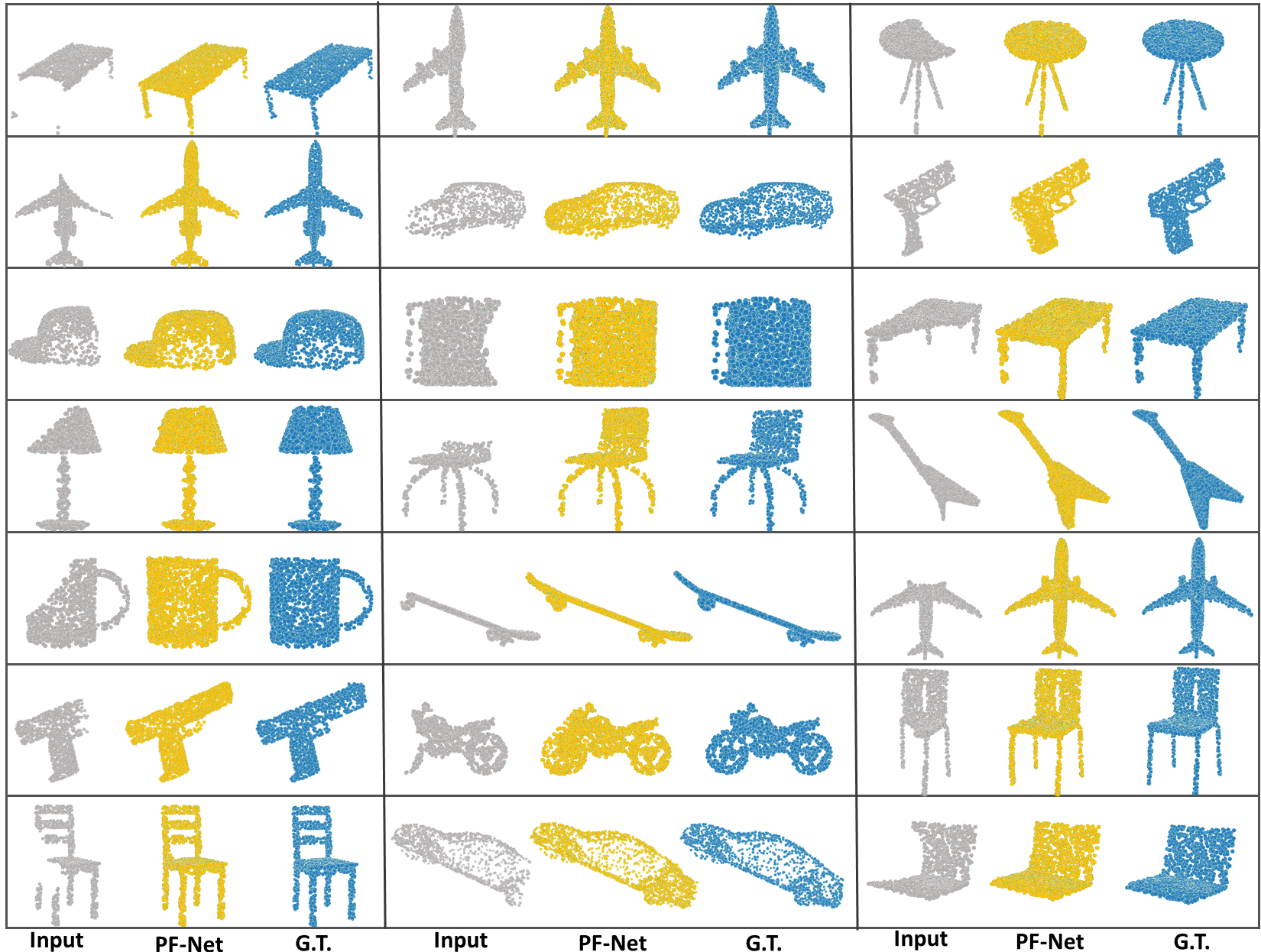}
\caption{More completion results of PF-Net. The gray point clouds are the input of our method. We combine the prediction of our method and the input incomplete point cloud to form the completion results(indicate by yellow). The blue point clouds are the ground truth. }  %图片的名称
%\fbox{\rule{0pt}{2in}
%\rule{.9\linewidth}{0pt}
\label{Fig.10}   %标签，用作引用
\end{figure*}
More completion results of our PF-Net are shown in Fig.\ref{Fig.10}, including airplane, table, chair, car, pistol, cap, mug, lamp, skateboard, motorbike, laptop, guitar. We combine the output of our PF-Net and the input incomplete point cloud to form the final completion  results.

\begin{thebibliography}{1}\itemsep=-1pt
\bibitem{achlioptas2018learning}
Panos Achlioptas, Olga Diamanti, Ioannis Mitliagkas, and Leonidas~J Guibas.
\newblock Learning representations and generative models for {3D} point clouds.
\newblock {\em ICML}, 2018.

\bibitem{birdal2017a}
Tolga Birdal and Slobodan Ilic.
\newblock A point sampling algorithm for {3D} matching of irregular geometries.
\newblock {\em IROS}, 2017.

\bibitem{cai2018cascade}
Zhaowei Cai and Nuno Vasconcelos.
\newblock Cascade {R-CNN}: Delving into high quality object detection.
\newblock {\em CVPR}, 2018.

\bibitem{chen2019hybrid}
Kai Chen, Jiangmiao Pang, Jiaqi Wang, Yu Xiong, Xiaoxiao Li, Shuyang Sun,
  Wansen Feng, Ziwei Liu, Jianping Shi, Wanli Ouyang, et~al.
\newblock Hybrid task cascade for instance segmentation.
\newblock {\em CVPR}, 2019.

\bibitem{dai2017shape}
Angela Dai, Charles~R Qi, and Matthias Niebner.
\newblock Shape completion using {3D-Encoder-Predictor CNNs} and shape
  synthesis.
\newblock {\em CVPR}, 2017.

\bibitem{Fan2016A}
Haoqiang Fan, Su Hao, and Leonidas Guibas.
\newblock A point set generation network for {3D} object reconstruction from a
  single image.
\newblock {\em CVPR}, 2017.

\bibitem{feng2019hypergraph}
Yifan Feng, Haoxuan You, Zizhao Zhang, Rongrong Ji, and Yue Gao.
\newblock Hypergraph neural networks.
\newblock {\em AAAI}, 2019.

\bibitem{gadelha2018multiresolution}
Matheus Gadelha, Rui Wang, and Subhransu Maji.
\newblock Multiresolution tree networks for {3D} point cloud processing.
\newblock {\em ECCV}, 2018.

\bibitem{goodfellow2014generative}
Ian Goodfellow, Jean Pougetabadie, Mehdi Mirza, Bing Xu, David Wardefarley,
  Sherjil Ozair, Aaron Courville, and Yoshua Bengio.
\newblock Generative adversarial nets.
\newblock {\em NeurIPS}, 2014.

\bibitem{Han2017High}
Xiaoguang Han, Li Zhen, and Huang Haibin.
\newblock High-resolution shape completion using deep neural networks for
  global structure and local geometry inference.
\newblock In {\em ICCV}, 2017.

\bibitem{he2018triplet-center}
Xinwei He, Yang Zhou, Zhichao Zhou, Song Bai, and Xiang Bai.
\newblock Triplet-center loss for multi-view {3D} object retrieval.
\newblock {\em CVPR}, 2018.

\bibitem{jaiswal2018capsulegan:}
Ayush Jaiswal, Wael Abdalmageed, Yue Wu, and Premkumar Natarajan.
\newblock {CapsuleGAN}: Generative adversarial capsule network.
\newblock {\em ECCV}, 2018.

\bibitem{kanezaki2018rotationnet:}
Asako Kanezaki, Yasuyuki Matsushita, and Yoshifumi Nishida.
\newblock {RotationNet}: Joint object categorization and pose estimation using
  multiviews from unsupervised viewpoints.
\newblock {\em CVPR}, 2018.

\bibitem{Kirillov2019Panoptic}
Alexander Kirillov, Ross Girshick, Kaiming He, and Piotr Dollár.
\newblock Panoptic feature pyramid networks.
\newblock {\em CVPR}, 2019.

\bibitem{li2018point}
Chunliang Li, Manzil Zaheer, Yang Zhang, Barnabas Poczos, and Ruslan
  Salakhutdinov.
\newblock Point cloud {GAN}.
\newblock {\em ICLR workshops}, 2018.

\bibitem{lin2018learning}
Chenhsuan Lin, Chen Kong, and Simon Lucey.
\newblock Learning efficient point cloud generation for dense {3D} object
  reconstruction.
\newblock {\em AAAI}, 2018.

\bibitem{Lin2017Feature}
Tsung~Yi Lin, Piotr Dollar, Ross Girshick, Kaiming He, Bharath Hariharan, and
  Serge Belongie.
\newblock Feature pyramid networks for object detection.
\newblock {\em CVPR}, 2017.

\bibitem{liu2019relation-shape}
Yongcheng Liu, Bin Fan, Shiming Xiang, and Chunhong Pan.
\newblock Relation-shape convolutional neural network for point cloud analysis.
\newblock {\em CVPR}, 2019.

\bibitem{pang2019libra}
Jiangmiao Pang, Kai Chen, Jianping Shi, Huajun Feng, Wanli Ouyang, and Dahua
  Lin.
\newblock Libra {R-CNN}: Towards balanced learning for object detection.
\newblock {\em CVPR}, 2019.

\bibitem{pathak2016context}
Deepak Pathak, Philipp Krahenbuhl, Jeff Donahue, Trevor Darrell, and Alexei~A
  Efros.
\newblock Context encoders: Feature learning by inpainting.
\newblock {\em CVPR}, 2016.

\bibitem{qi2016pointnet:}
Charles~R Qi, Hao Su, Kaichun Mo, and Leonidas~J Guibas.
\newblock Pointnet: Deep learning on point sets for {3D} classification and
  segmentation.
\newblock {\em CVPR}, 2017.

\bibitem{qi2017pointnet++:}
Charles~R Qi, Li Yi, Hao Su, and Leonidas~J Guibas.
\newblock Pointnet++: Deep hierarchical feature learning on point sets in a
  metric space.
\newblock {\em NeurIPS}, 2017.

\bibitem{sabour2017dynamic}
Sara Sabour, Nicholas Frosst, and Geoffrey~E Hinton.
\newblock Dynamic routing between capsules.
\newblock {\em NeurIPS}, 2017.

\bibitem{sarmad2019rl-gan-net:}
Muhammad Sarmad, Hyunjoo Lee, and Young~Min Kim.
\newblock {RL-GAN-Net}: A reinforcement learning agent controlled {GAN} network
  for real-time point cloud shape completion.
\newblock {\em CVPR}, 2019.

\bibitem{tian2019fcos:}
Zhi Tian, Chunhua Shen, Hao Chen, and Tong He.
\newblock Fcos: Fully convolutional one-stage object detection.
\newblock {\em CVPR}, 2019.

\bibitem{wang2018an}
Dilin Wang and Qiang Liu.
\newblock An optimization view on dynamic routing between capsules.
\newblock {\em ICLA}, 2018.

\bibitem{wang2017shape}
Weiyue Wang, Qiangui Huang, Suya You, Chao Yang, and Ulrich Neumann.
\newblock Shape inpainting using {3D} generative adversarial network and
  recurrent convolutional networks.
\newblock {\em ICCV}, 2017.

\bibitem{wu2016learning}
Jiajun Wu, Chengkai Zhang, Tianfan Xue, William~T Freeman, and Joshua~B
  Tenenbaum.
\newblock Learning a probabilistic latent space of object shapes via {3D}
  generative-adversarial modeling.
\newblock {\em NeurIPS}, 2016.

\bibitem{wu20153d}
Zhirong Wu, Shuran Song, Aditya Khosla, Fisher Yu, Linguang Zhang, Xiaoou Tang,
  and Jianxiong Xiao.
\newblock {3D} shapenets: A deep representation for volumetric shapes.
\newblock {\em CVPR}, 2015.

\bibitem{yang20173d}
Bo Yang, Hongkai Wen, Sen Wang, Ronald Clark, Andrew Markham, and Niki Trigoni.
\newblock {3D} object reconstruction from a single depth view with adversarial
  learning.
\newblock {\em ICCV Workshops}, 2017.

\bibitem{yang2017foldingnet:}
Yaoqing Yang, Chen Feng, Yiru Shen, and Dong Tian.
\newblock Foldingnet: Point cloud auto-encoder via deep grid deformation.
\newblock {\em CVPR}, 2018.

\bibitem{yi2016a}
Li Yi, Vladimir~G Kim, Duygu Ceylan, Ichao Shen, Mengyan Yan, Hao Su, Cewu Lu,
  Qixing Huang, Alla Sheffer, and Leonidas~J Guibas.
\newblock A scalable active framework for region annotation in {3D} shape
  collections.
\newblock {\em ACM Trans. on Graphics}, 35(6):210:1--210:12, 2016.

\bibitem{yu2018pu-net:}
Lequan Yu, Xianzhi Li, Chiwing Fu, Daniel Cohenor, and Phengann Heng.
\newblock {PU-Net}: Point cloud upsampling network.
\newblock {\em CVPR}, 2018.

\bibitem{yu2018multi-view}
Tan Yu, Jingjing Meng, and Junsong Yuan.
\newblock Multi-view harmonized bilinear network for {3D} object recognition.
\newblock {\em CVPR}, 2018.

\bibitem{yuan2018pcn:}
Wentao Yuan, Tejas Khot, David Held, Christoph Mertz, and Martial Hebert.
\newblock {PCN}: Point completion network.
\newblock {\em 3DV}, 2018.

\bibitem{zhao20183d}
Yongheng Zhao, Tolga Birdal, Haowen Deng, and Federico Tombari.
\newblock {3D} point capsule networks.
\newblock {\em CVPR}, 2018.

\bibitem{zhou2018voxelnet:}
Yin Zhou and Oncel Tuzel.
\newblock {VoxelNet}: End-to-end learning for point cloud based {3D} object
  detection.
\newblock {\em CVPR}, 2018.
\end{thebibliography}
\end{document}